\documentclass[english]{cccconf}
\usepackage[comma,numbers,square,sort&compress]{natbib}
\usepackage{epstopdf}

\usepackage{amsmath}
\usepackage{algorithm}
\usepackage{algorithmic}
\begin{document}

\title{Deep Reinforcement Learning Based High-level Driving Behavior Decision-making Model in Heterogeneous Traffic}

\author{Zhengwei Bai \aref{bjtu},
        Baigen Cai \aref{bjtu,lab},
        Wei Shangguan* \aref{bjtu,lab},
        Linguo Chai \aref{bjtu,lab}}


\affiliation[bjtu]{School of Electronic and Information Engineering, Beijing Jiaotong University, Beijing 100044, P.~R.~China}
\affiliation[lab]{State Key Laboratory of Rail Traffic Control and Safety, Beijing Jiaotong University, Beijing 100044, P.~R.~China
        \email{zwbai@bjtu.edu.cn, wshg@bjtu.edu.cn}}

\maketitle

\begin{abstract}
High-level driving behavior decision-making is an open-challenging problem for connected vehicle technology, especially in heterogeneous traffic scenarios. In this paper, a deep reinforcement learning based high-level driving behavior decision-making approach is proposed for connected vehicle in heterogeneous traffic situations. The model is composed of three main parts: a data preprocessor that maps hybrid data into a data format called hyper-grid matrix, a two-stream deep neural network that extracts the hidden features, and a deep reinforcement learning network that learns the optimal policy. Moreover, a simulation environment, which includes different heterogeneous traffic scenarios, is built to train and test the proposed method. The results demonstrate that the model has the capability to learn the optimal high-level driving policy such as driving fast through heterogeneous traffic without unnecessary lane changes. Furthermore, two separate models are used to compare with the proposed model, and the performances are analyzed in detail. 
\end{abstract}


\footnotetext{*Corresponding author \\
This work is supported by National Key Research and Development Program of China (2018YFB1600604, 2016YFB1200100), National Natural Science Foundation of China Monumental Projects (61490705, 61773049).}

\section{Introduction}

Connected vehicle (CV) technology is capable of providing efficient and safe driving by perceiving comprehensive environmental information through the vehicle to everything (V2X) communication. Considering, however, that the transition period when CV share the roadway with human-driven vehicles may last several decades, it is crucial to ensure the CV could run efficiently in heterogeneous traffic situations.

For the development of CV technology, one of the crucial topics on autonomous driving is to make the optimal driving policy or supervisor that (1) provides high efficient driving strategies to improve traffic efficiency, and (2) ensures safety during the driving process under complex traffic conditions.

The high-level driving policy aims at making the appropriate decision, i.e., choosing driving behaviors to achieve some optimal purpose such as driving through dense traffic at a higher speed.

In order to make the high-level driving policy have the capacity to handle both efficiency and safety in such a heterogeneous traffic scenario, it is critical to propose a high-level driving behavior decision-making model.

In conventional research on the driving strategies of the CV technology, there are two main approaches: (1) model-based framework which contains information perception, path planning, motion planning, and motion control and (2) learning-based framework such as deep learning based or deep RL based. Although these state-of-art rule-based methods\cite{cheng2005} can generate precise path and control in specific environments, the complexity and variety in real life road traffic environment limit the traditional method application. Thus, the learning-based algorithms are widely exploited in recent researches, since it can handle complex environments with tremendous power \cite{Mnih2015Human}\cite{Meixin}.

An active area of the research is deep learning based approaches.  Bojarski \cite{Bojarski2016End} proposed an end-to-end approach based on the deep convolutional neural network to train the network map raw pixels from a single front-facing camera directly to steering commands. Baidu \cite{Yu2017Baidu} proposed a convolutional neural network (CNN) and convolutional long short-term memory based end-to-end reacting control model, which includes lateral control and longitudinal control. Al-Qizwini \cite{Al2017Deep} proposed a deep learning based direct perception approach for autonomous driving using GoogLeNet (GLAD) which makes no unrealistic assumptions about the autonomous vehicle or its surroundings using deep learning.
For the reinforcement learning based methods, Zuo \cite{Zuo2017Continuous} proposed a continuous reinforcement learning method which integrates Deep Deterministic Policy Gradient with human demonstrations. The algorithm can learn more demonstrator`s preferences and accelerate the training process at the same time.
Fridman \cite{Fridman2018DeepTraffic} created a simulation where one of the vehicles is a reinforcement learning agent operating according to the Q-function estimated by a neural network. The results demonstrate that the reinforcement learning algorithm can be remarkably powerful for driving policy learning.
Min \cite{Min} proposed an autonomous driving framework using deep Q-learning to determine advanced driver assistance system (ADAS) functions known as high level driving policy determination.

However, the above approaches have three main limitations: (1) some models learned the driving strategy by outputting the proper steering angle and amount of throttle and brake directly, which means the proposed driving policy will change every time the parameters of the vehicle changed; (2) some models assumes that all the other vehicles are the non-intelligent vehicle and have the same driving behavior policy which is obviously not realistic in real-life traffic situations; and (3) they used the raw data, which perceived from the simulation environment, as the input of the deep neural network directly. Although these frameworks worked well in their simulation environment, using simulated raw data as input directly will limit the generalization capacity of the model, because the real-life traffic environments are far from the same as these simulated ones.

In this paper, we proposed (1) a deep RL-based high-level driving behavior decision-making model for the CV in heterogenous dense traffic situations; (2) a data mapping algorithm to transform the raw V2X information into a unified data format, named as hyper-grid matrix (HGM), which decoupled the raw V2X data and the neural network; and (3) a parameter-sharing two-stream deep neural network to cope with the hybrid input data.
\begin{figure}[!htb]
  \centering
  \includegraphics[width=0.5\textwidth]{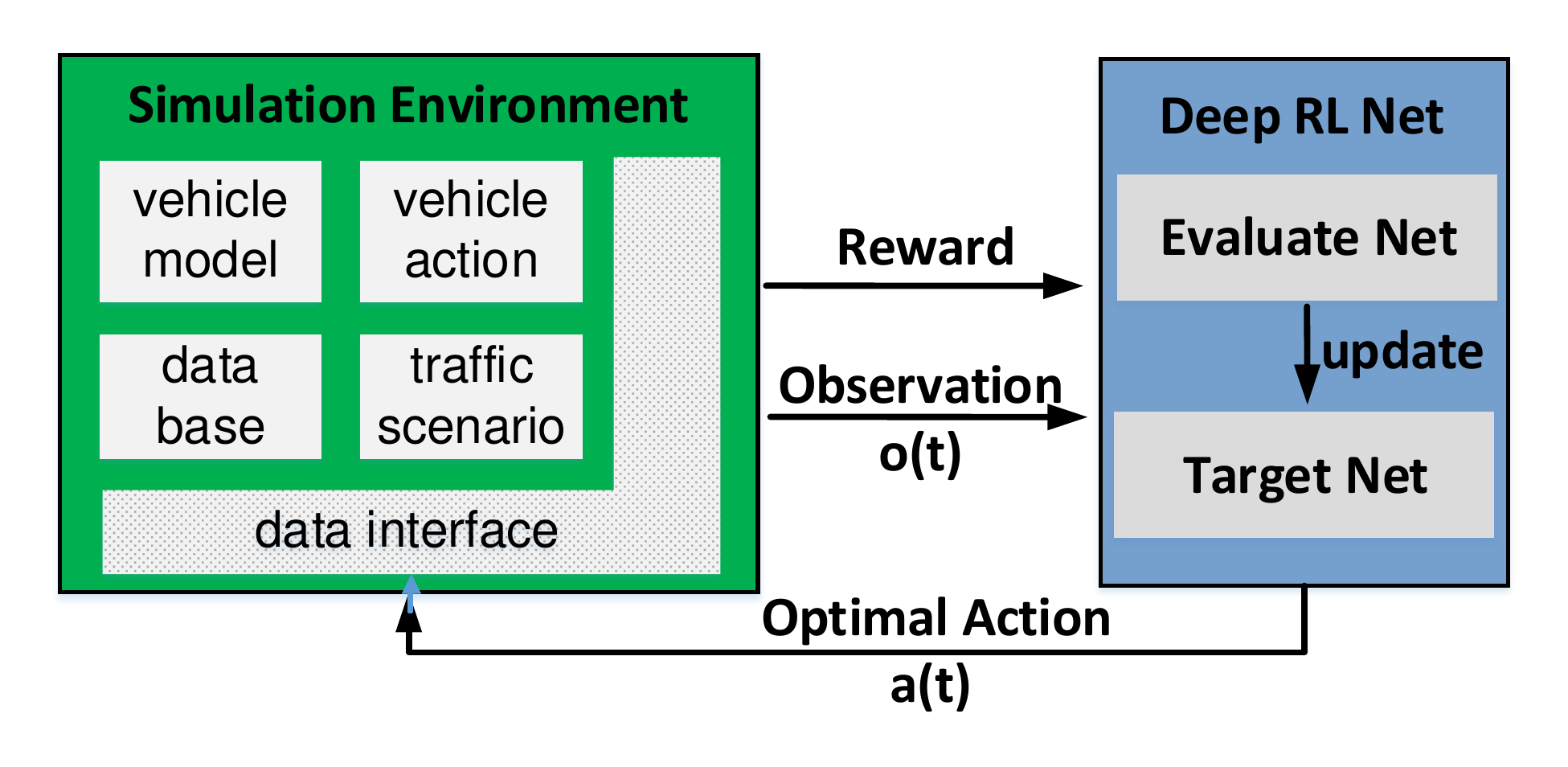}
  \caption{The schematic diagram of the deep RL based high-level driving behavior decision-making model.}
  \label{framework}
\end{figure}

Fig.\ref{framework} shows a schematic diagram of the proposed high-level driving behavior decision-making model based on deep RL. The learning process consists of two phases: the observing phase and the training phase. During the observing phase, data are collected and stored in the database as memory data. During the training phase, these data are fed into a simulation environment where the agent learns how to take the optimal action (high-level driving behavior) based on the observation state (V2X data and raw image) and reward function. Through these interactions, an optimal policy, or high-level driving behavior decision-making model is generated, i.e., deciding the appropriate high-level driving behavior to achieve the driving purpose. The model, or policy, can be continuously updated when more data are fed in. This optimal policy will act as the executing policy in the CV driving process. In order to train the proposed model and test its performance, a simulation environment is developed. The performance of the proposed model is compared with two different models and tested in heterogeneous traffic scenarios.

This paper begins with a description of deep reinforcement learning algorithms. The simulation traffic scenario and proposed deep RL based high-level driving behavior decision-making model are then specified, followed by training and performance evaluation of the model. The final section is devoted to the conclusion.

\section{Deep reinforcement learning}
Deep reinforcement learning (DRL) combines deep learning (DL) with reinforcement learning (RL) to learn control strategies directly from high-dimensional raw data. With Alpha Go \cite{Silver2016Mastering} defeating the strongest human players, reinforcement learning demonstrates its remarkable capacity for policy learning. With the proliferation for deep RL algorithm, an increasing number of deep RL algorithms, such as deep Q network (DQN) \cite{Mnih2015Human}, Double DQN \cite{Hasselt2015Deep}, Dueling DQN \cite{Wang2015Dueling}, etc., have been proposed and applied in various studies \cite{Arulkumaran2017Deep}. In addition, the algorithms are briefly explained as follows.

\subsection{DQN}
Deep Q Network is a combination between the convolutional neural network (CNN) and Q-Learning algorithm. The input of CNN is the original image data (as observation state O), and the output is the evaluation value (Q-value) corresponding to each action in state A. Then, according to the e-greedy algorithm, an action is selected from the action space. After the execution of action $A_{t}$, a reward $R_{t}$ and an observation state $O_{t+1}$ can be get from the environment.

\subsection{Double DQN}
Although DQN has been applied for many applications, it has a problem of overestimating. Thus, an optimal DQN architecture named double DQN (DDQN) was developed to optimize the overestimation problem. For conventional DQN, both selecting and evaluating network use the same Q-function which cause the overestimation of action value. In order to optimize the performance limited by overestimation, DDQN constructed a novel network architecture which includes two Q-functions for selecting and evaluating action.

\subsection{Dueling DQN}
In many visual perception-based DRL tasks, the value functions of different state actions are disparate, but in some states, the size of the value function is independent of the action. Thus, Dueling DQN is constructed with two streams which separately estimate (scalar) state-value and the advantages of each action and shows significant performance improvement than DQN.
        
\subsection{Selected Algorithm}
In this study, considering that the high-level driving behavior decision-making is a discrete output problem, Dueling DQN and prioritized experience replay \cite{Schaul2015Prioritized} algorithms are applied to build our proposed model because of their remarkable performance in learning efficiency and robustness.
The equation for calculating Q-value of Dueling DQN, shown as follow, is designed to aggregate the states-value and action advantages.
\begin{equation}
\label{Dueling}
\begin{split}
   Q( S_{t}, A_{t}; \theta, \alpha, \beta) & = V(S_{t}; \theta, \beta) + A(S_{t}, A_{t}; \theta, \alpha) \\
     & -\frac{1}{|A|} \sum_{A_{t}}A(S_{t}, A_{t}; \theta, \alpha)
\end{split}
\end{equation}

As is shown in the equation, $\alpha$ represents the parameters of $A$ (the advantage function). Besides, $\beta$ represents the parameters of $V$ (the state-value function) and $\theta$ is parameters of neural network.

\section{Approach to The Proposed Model}

\subsection{Simulation Setup}
\begin{figure}[!htb]
  \centering
  \includegraphics[width=0.4\textwidth]{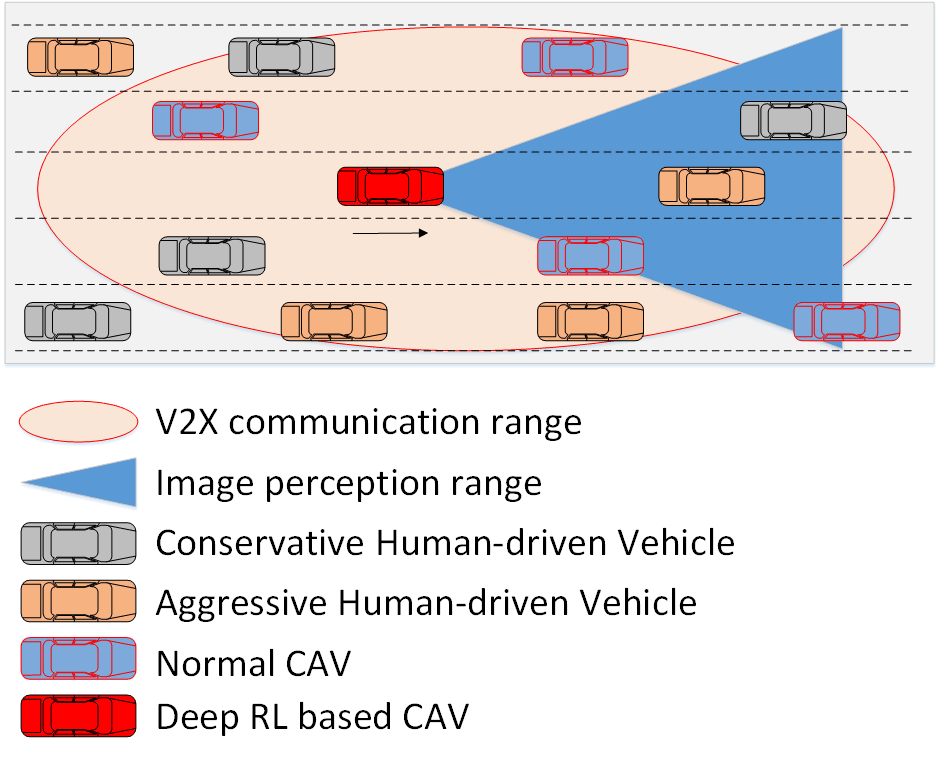}
  \caption{The description of heterogeneous traffic scenario.}
  \label{trafficscenario}
\end{figure}

To train and evaluate the proposed approach, we constructed a simulation platform which includes several traffic scenarios. The simulation environment, developed by Unity ML-Agent framework, is heterogeneous traffic scenarios, i.e., a host vehicle (HV) driving through the dense traffic in a five-lane highway road situation which is shown in Fig.\ref{trafficscenario}. For the dense traffic flow, all vehicles have five high-level driving behaviors or known as actions which include acceleration, deceleration, lane change to the left lane, lane change to the right lane, no action.

In the considering of the characteristics of heterogeneous traffic, the perceptive patterns of the deep RL based HV are set as follows: (1) the HV could get data from other CV through V2X communication, and (2) there is a front camera that can get the general image data. There are four kinds of vehicles which are the conservative human-driven vehicle (CHV), aggressive human-driven vehicle (AHV), normal CV (NCV), and deep RL based CV separately (Host Vehicle). The vehicle model is shown in Table~\ref{vehiclemodel}.

\subsection{Observation State}
\begin{figure}[!htb]
  \centering
  \includegraphics[width=0.5\textwidth]{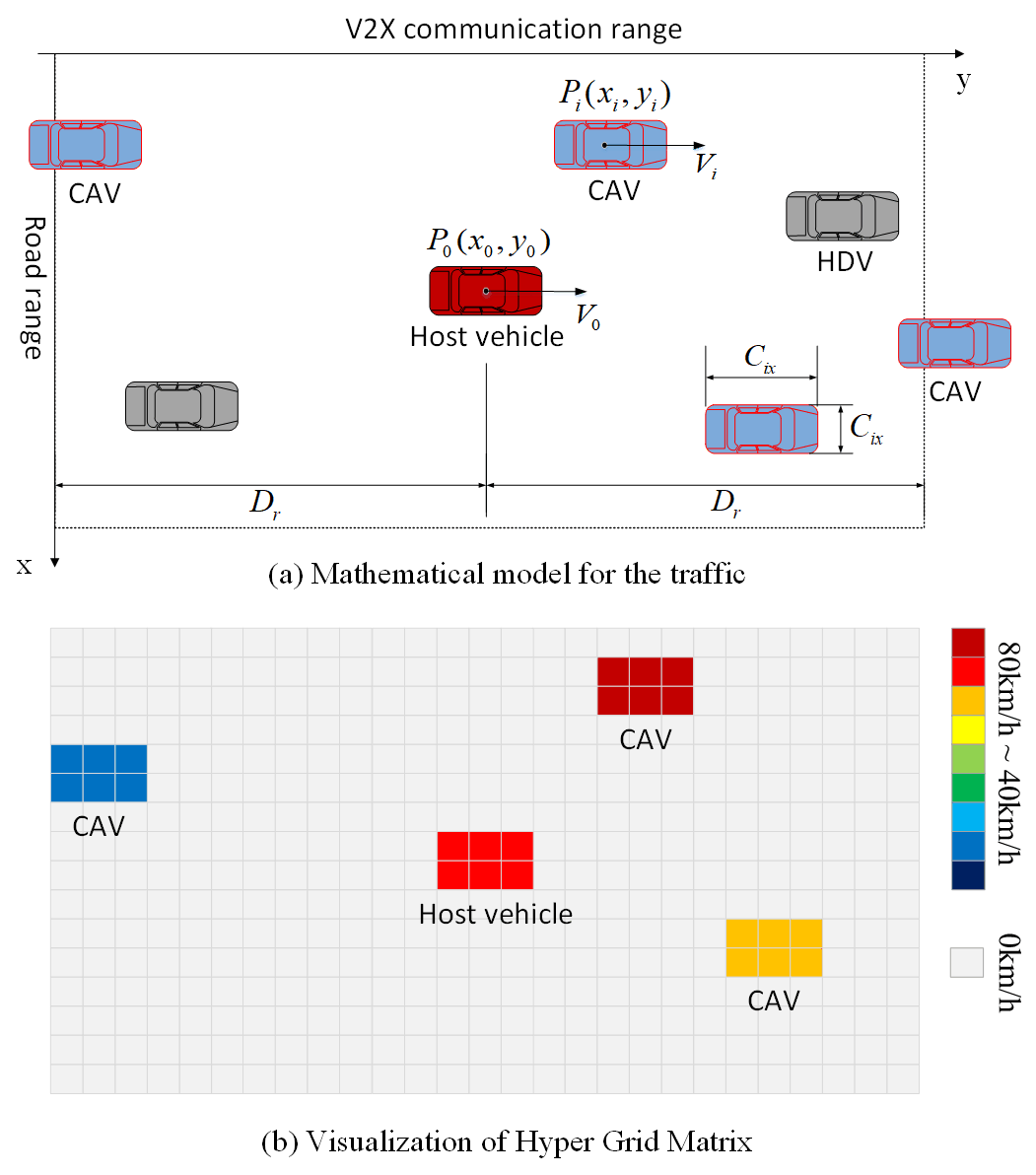}
  \caption{The visualization of the HDM mapping process.}
  \label{hypergird}
\end{figure}

For conventional state-of-art reinforcement learning based driving policy learning model, most of the observation states are constructed by raw data collected from the simulation environment. In this paper, we proposed a new data format called Hyper-Grid Matrix (HGM) which mapped the raw V2X data into a high dimensional unified data format, shown as Fig.\ref{hypergird}.
For environment perception, the red vehicle is set as the HV (the agent) which can communicate with other CV through V2X communication. The V2X data includes the current speed $V_{t}$, absolute position $P_{i}(x, y)$ and collision volume of the vehicle $(c_{ix}, c_{iy})$ where $i$ means the number of the vehicle. The perception range is set to $D_{r}$ centered on the HV. In this case, the $ D_{r}$ is set to 40 meters and the $(c_{ix}, c_{iy})$ is set by the Unity Model.

The mapping algorithm from V2X data to the HGM is shown as Algorithm~\ref{alg:Framwork}. Firstly, the absolute coordinates of the vehicle are converted to relative coordinates based on the position of the host vehicle. Secondly, the mapping position and size of the vehicle in the hyper grid are determined based on the relative position and collision volume of the vehicle. Finally, the value of the HDM is set to the normalized value of the vehicle speed.

\begin{algorithm}[!htb]
\caption{ Framework of hyper grid matrix mapping.}
\label{alg:Framwork}
\begin{algorithmic}[1] 
\REQUIRE ~~\\ 
The V2X communication range $D_r$;\\
The CAV data: position $P_{i}(x_{i},y_{i})$, speed $v_{i}$, collision size $c_{ix}$,$c_{iy}$;
\ENSURE ~~\\ 
The hyper grid matrix, $HDM[m][n]$;
\STATE Set the V2X communication range, $D_r$;
\label{ code:fram:extract }
\STATE Initialize the matrix, $HDM[m][n]$ = 0;
\label{code:fram:trainbase}
\STATE Initialize the position of host vehicle, $P(x_{0}, y_{0})$
\label{code:fram:add}
\IF{other CAV is within the V2X communication range, $|y_{i}-y_{0}| < D_r$;}
\STATE Receive the CAV data $P_{i}(x_{i},y_{i})$, speed $v_{i}$, collision size $c_{ix}$,$c_{iy}$
\ENDIF
\label{code:fram:classify}
\IF{$(x_{i}-c_{ix}) < m < (x_{i} + c_{ix})$ and $(y_{i}-c_{iy}) < n < (y_{i} + c_{iy})$}
\STATE $HDM[m][n]$ = $\frac{v - (v_{max} + v_{min})/2 }{(v_{max} + v_{min})/2}$
\ENDIF
\label{code:fram:select}
\RETURN $HDM[m][n]$; 
\end{algorithmic}
\end{algorithm}

\subsection{Vehicle Action Space}

In this paper, the primary purpose is to find an optimal policy to generate the appropriate high-level driving behavior for the CAV to achieve efficiency and safety. Thus, for the action space which is known as the high-level driving behavior space, we defined the action space based on the driving characteristics of the dense traffic. There are 5 actions are built in total which are (1)Acceleration, (2)Deceleration, (3)Change lane to the left, (4) Change lane to the right, and (5)Take no action. The descriptions about the action space and vehicle dynamics are defined as follow:
  \begin{equation}
    a_{acc} = -a_{dec} = \left\{
    \begin{array}{rcl}
    a_{lon}, &{v_{min} \leq v \leq v_{max} }\\
    0, & {v_{t} = v_{max} }
    \end{array} \right.
  \end{equation}
  \begin{equation}
    v_{toRight} = -v_{toLeft} = \left\{
    \begin{array}{rcl}
    v_{lat}, &{moving}\\
    0, & { finished }
    \end{array} \right.
  \end{equation}
where $a_{acc}$, $a_{dec}$, $v_{toRight}$, $v_{toLeft}$, $a_{lon}$, $v_{lat}$ represent the value of acceleration, deceleration, speed of change lane to right, speed of change lane to left, longitude acceleration, and latitude velocity separately. Moreover, the action model of all kinds of vehicles are defined as Table~\ref{vehiclemodel}, where $r_{lon}$ and $r_{lat}$ represent the ratios of choosing the longitude and latitude actions randomly.
\begin{table}[!htb]
\centering
\caption{The model of all kinds of vehicle}
\label{vehiclemodel}
\begin{tabular}{l|l|l|l|l}
\hhline
			Type & $a_{lon}$ & $v_{lat}$ & $r_{lon}$ & $r_{lat}$ \\ \hline
			CHV&  2$m/s^{2}$ & 1$m/s$    & 0.2       &   0.2     \\
			AHV&  4$m/s^{2}$ & 2$m/s$    & 0.4       &   0.4     \\
			NCV&  2$m/s^{2}$ & 1$m/s$    & 0.2       &   0.2     \\
			HV	&  2$m/s^{2}$ & 1$m/s$    & 0.0       &   0.0     \\
\hhline
\end{tabular}
\end{table}
\begin{figure*}[htbp]
  \centering
  \includegraphics[width=1\textwidth]{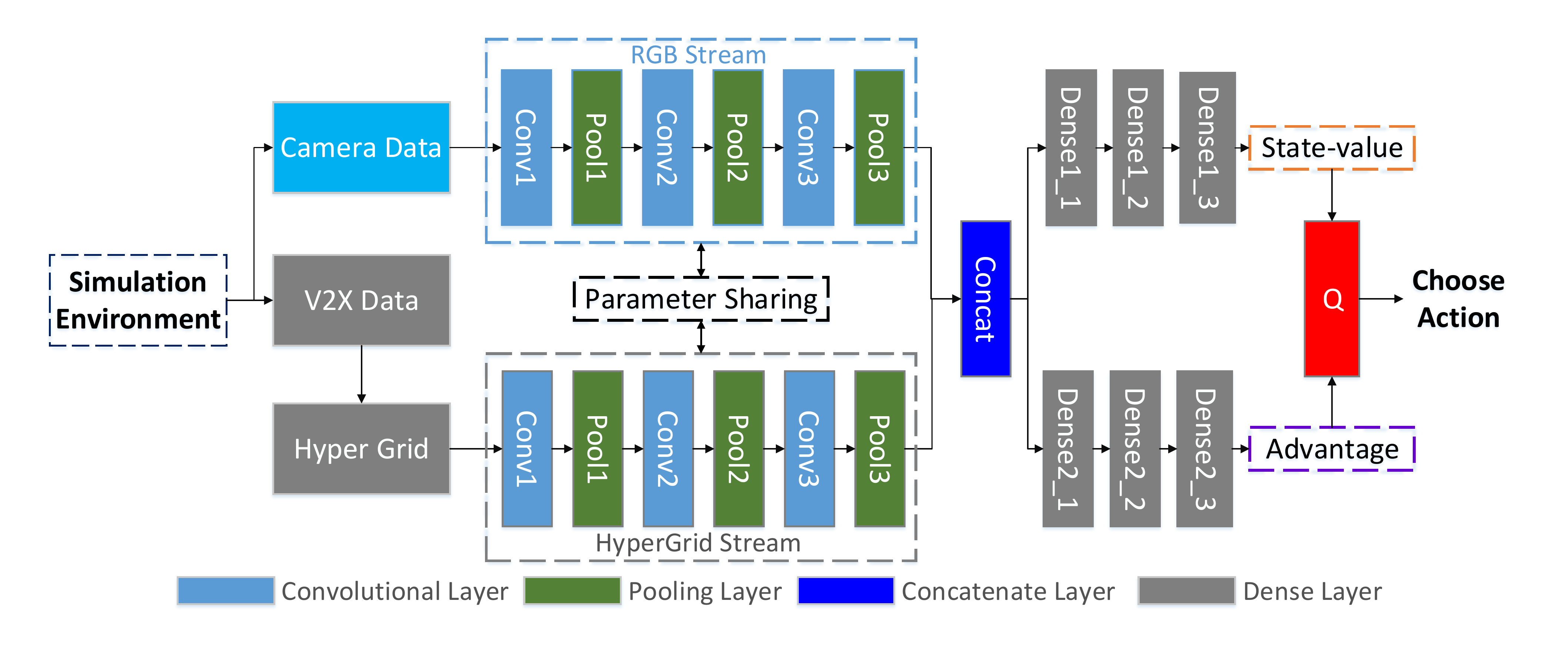}
  \caption{Architecture of the proposed deep neural network.}
  \label{MainNet}
\end{figure*}
\subsection{Reward Function}
In order to make the autonomous driving policy and human driving habits more compatible, some principles of reward based on human driving behavior habits were considered: (1) Avoid any collisions, (2) Prevent dangerous action which may end up with a hazardous saturation, (3) Change lanes as little as possible, (4) Get higher speed, and (5) Overtake more vehicles. Based on these principles, the reward function is designed as Algorithm~\ref{alg:reward}, 
\begin{algorithm}[!htb]
\caption{ Reward Function Descriptions.}
\label{alg:reward}
\begin{algorithmic}[1] 
\REQUIRE ~~\\ 
The state of host vehicle $S(t)$;
\ENSURE ~~\\ 
The reward value $R_{all}(t)$;
\STATE Initialization: $R_{all} = R_{v}=R_{c}=R_{d}=R_{l}=R_{o}=0$

\STATE $R_{v}(v_{t}) = (1 + \frac{ v_{t} - v_{min}}{v_{max} - v_{min}})^ {2} - 1$;
\label{ code:fram:extract }

\IF{Collision happens;}
\STATE $R_{c} = R_{collision}$
\ENDIF
\label{code:fram:classify}

\IF{Dangerous action happens;}
\STATE $R_{d} = R_{dangerous}$
\ENDIF
\label{code:fram:classify}

\IF{Lane-change action happens;}
\STATE $R_{l} = R_{lanechange}$
\ENDIF
\label{code:fram:classify}

\IF{Host vehicle overtakes another;}
\STATE $R_{o} = R_{overtake}$
\ENDIF
\label{code:fram:classify}

\STATE $ R_{all} =    R_{v}(v_{t}) + R_{c} + R_{d} + R_{lc} + R_{o}$
\label{code:fram:classify}

\RETURN $R_{all}$; 
\end{algorithmic}
\end{algorithm}
where $v_{t}$ represents the current speed of host vehicle, the $v_{max}$ is the maximum speed specified and the $v_{min}$ is the minimum speed specified. $R_{v}(v_{t})$ and $R_{overtake}$ represent the reward for speed and overtaking other vehicles respectively. In addition, $R_{collision}$, $R_{dangerous}$, $R_{lanechange}$ represent the penalty for host vehicle collision, making dangerous actions and lane changing respectively. Finally, $R_{all}$ means the whole reward at time $t$ with action $a$. The parameters of reward are given in Table~\ref{reward}

\begin{table}[!htb]
\centering
\caption{Parameters for Reward}
\label{reward}
\begin{tabular}{l|l|l|l|l|l|l}
\hhline
			Reward	&	$v_{max}$	&	$v_{min}$	&$r_{c}$	&	$r_{d}$	&	$r_{l}$	&	$r_{o}$	\\ \hline
			Parameters&	80$km/h$	&	40$km/h$	&	-20				&	-2					&	-1					&	 5	\\

\hhline
\end{tabular}
\end{table}

\subsection{Network Architecture}

The network architecture of the proposed approach for high-level driving behavior decision-making is shown as Fig.\ref{MainNet}. The network consists of three parts: (1) getting raw image data and transforming the V2X data into HGM; (2) the camera data and HGM data will be fed into a two-stream deep neural networks whose parameters are shared; and (3) the RGB stream and Hyper Grid stream will be concatenated together and then fed into two separate full connected neural networks (FCNN) to get the state value and advantage value which consist the Q value to choose action (the high-level driving behavior); The parameters of the neural network are shown as Table~\ref{netparameters}.
\begin{table}[!htb]
\centering
\caption{Parameters for deep neural network}
\label{netparameters}
\begin{tabular}{l|l|l|l|l|l}
\hhline
			Layer			&	Actuation	&	Patch size	 &Stride& Filter & Unit		\\ \hline
			Conv1			&	ReLU		&	8*8			 & 4 	 & 32 & -		\\ 
			Conv2			&	ReLU		&	4*4			 & 2 	 & 64 &-		\\ 
			Conv3			&	ReLU		&	3*3		 	& 1	 	& 64 & -		\\ 
			Dense1/2\_1	&	ReLU		&	- 			 & - 	 & -    & 512		\\ 
			Dense1/2\_2	&	ReLU		&	- 			 & - 	 & -    & 256		\\ 
			Dense1/2\_3	&	ReLU		&	- 			 & - 	 & -    & 1		\\ 

\hhline
\end{tabular}
\end{table}
Furthermore, in the first part, multi-frame HGM and image data was stacked into a higher dimensional data format base on time series. Thus the stacked data could contain both spatial and temporal information. Considering the correspondence of RGB image and HGM, two streams should interact with each other. Thus, the convolutional parameters of two streams are shared and updated simultaneously. Finally, the concatenated data was fed into two separate FCNNs built by the structure of Dueling DQN. For the two streams of FCNN, one was used to generate the state-value, and the other is used to generate the action-advantage. Finally, the state value and advantage are combined to create the Q value which is used to determine the optimal action.

\subsection{Network Update and Hyperparameters}

At each learning step, the weight coefficients of the proposed network were updated using the adaptive learning rate trick Adam\cite{Kingma2014Adam} in order to minimize the loss function .

The network parameters were updated as follows: considering the current state is s(t), The evaluation network can predict the Q(t) value of different actions corresponding to the current state. Then the greed policy was used to select the action with the largest Q value for the state transition. Through the target network, the Q values at time t+1 was generated and calculate the loss and then the evaluation network was updated.

The adopted hyperpatameters, parameters whose value are set prior to the beginning of the learning process, are shown in Table\ref{Hyperparameters}.

\begin{table}[!htb]
\centering
\caption{Hyperparameters and correspofing descriptions.}
\label{Hyperparameters}
\begin{tabular}{l|l|l}
\hhline
			Hyperparameters & Value & Description\\ \hline
			Learning rate $\alpha$ & 0.0025 &Learning rate used by Adam \\
			Discount factor $\gamma$& 0.99	&Q-learning discount factor \\
			Minibatch size	& 64& Number of training cases \\
			Observatioin step & 10000 & Steps used for observation \\
			Replay memory size& 10000 & Training cases in replay memory\\
			Target update steps& 10000 &Steps for target network update \\
			Traning steps& 1M & Steps for the training phase\\
			Testing steps& 200000 & Steps for the testing phase\\
\hhline
\end{tabular}
\end{table}

\section{Training and Test}
\subsection{Model Traning}
For the model training platform, the experimental environment is equipped with Intel(R) Core(TM) i7-6700 CPU @ 3.4GHz, 32 GB RAM, and an NVIDIA GTX 980 GPU. For the external brain of unity ML-Agent, the deep reinforcement learning neural networks were built by Tensorflow with Python. For the heterogeneous traffic scenario, the ratio of CV, AHV, CHV were set at 0.5, 0.25, and 0.25 separately. In the training process, hyperparameters are set as Table~\ref{Hyperparameters} and the simple exploration policy used in an $\varepsilon$-greedy policy with the $\varepsilon$ decreasing linearly from 1 to 0.1 over 1M steps.
\subsection{Test Procedure}
The proposed high-level driving behavior decision-making model was tested under the heterogeneous traffic flow with different ratio of CV and human-driving vehicles. There were 5 simulation environments were built to test the capacity of the proposed model, and the details of these simulation environments are shown in Table\ref{scenario}.
In order to demonstrate the performance of the proposed method, we compared the proposed method with Min’s method \cite{Min} and Coordinate method. In the Coordinate method, the HGM was not applied in the full algorithm, i.e., the V2X data was fed into the model directly. Thus, the capacity of HGM will be demonstrated through the comparison between the proposed method and the Coordinate method.
\begin{table}[!htb]
\centering
\caption{Test descriptions of heterogeneous traffic scenarios}
\label{scenario}
\begin{tabular}{l|l|l|l|l|l}
\hhline
Scenario NO. & 1 & 2 & 3 & 4 & 5\\ \hline
The ratio of CV &0.00 &0.25&0.50&0.75&1.00 \\
The ratio of AHV&0.50&0.375&0.25&0.125&0.00 \\
The ratio of CHV&0.50&0.375&0.25&0.125&0.00 \\
\hhline
\end{tabular}
\end{table}

\section{Results and Analysis}
\begin{figure}[!h]
  \centering
  \includegraphics[width=\hsize]{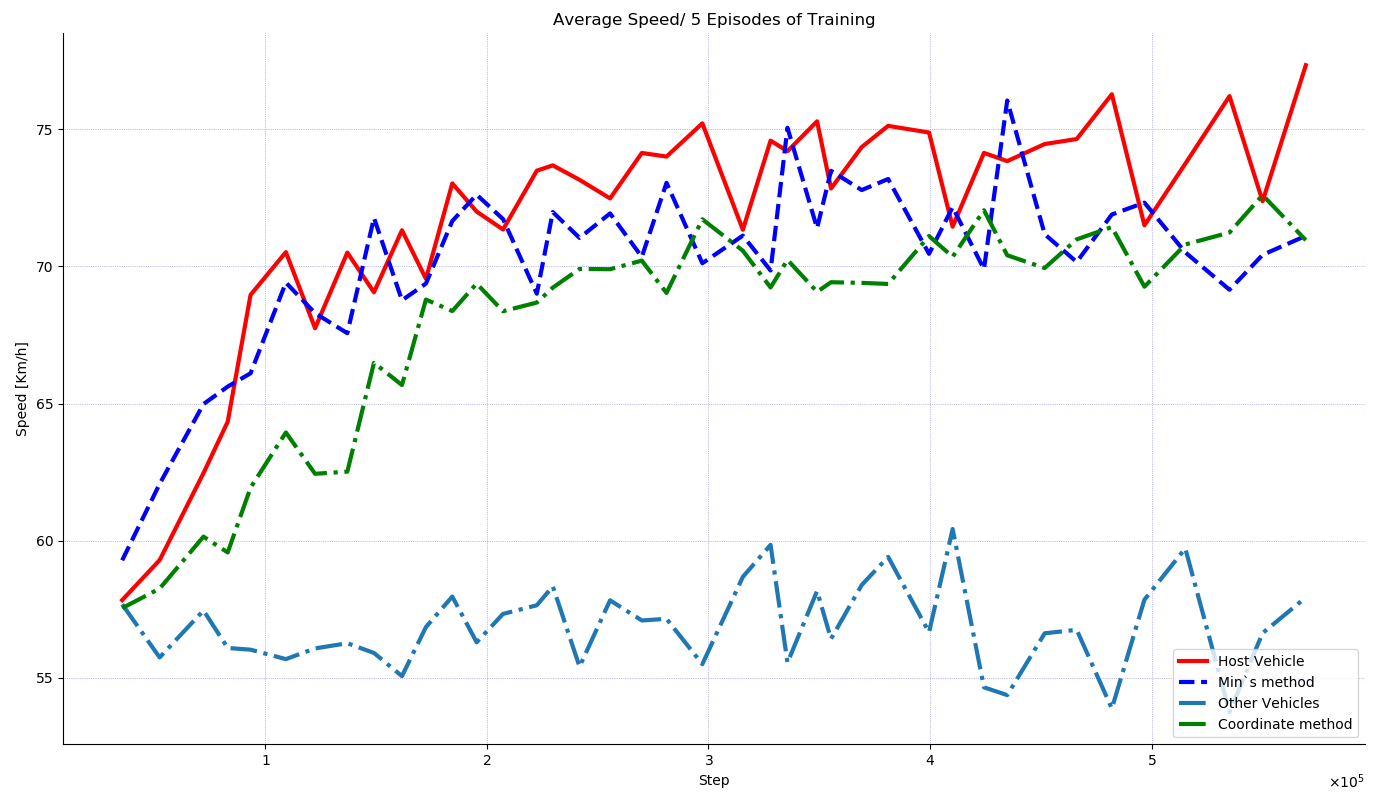}
  \caption{Average speed by proposed method, Min's method, Coordinate method, and Human-driven vehicle.}
  \label{speedComp}
\end{figure}

\begin{figure}[!h]
  \centering
  \includegraphics[width=\hsize]{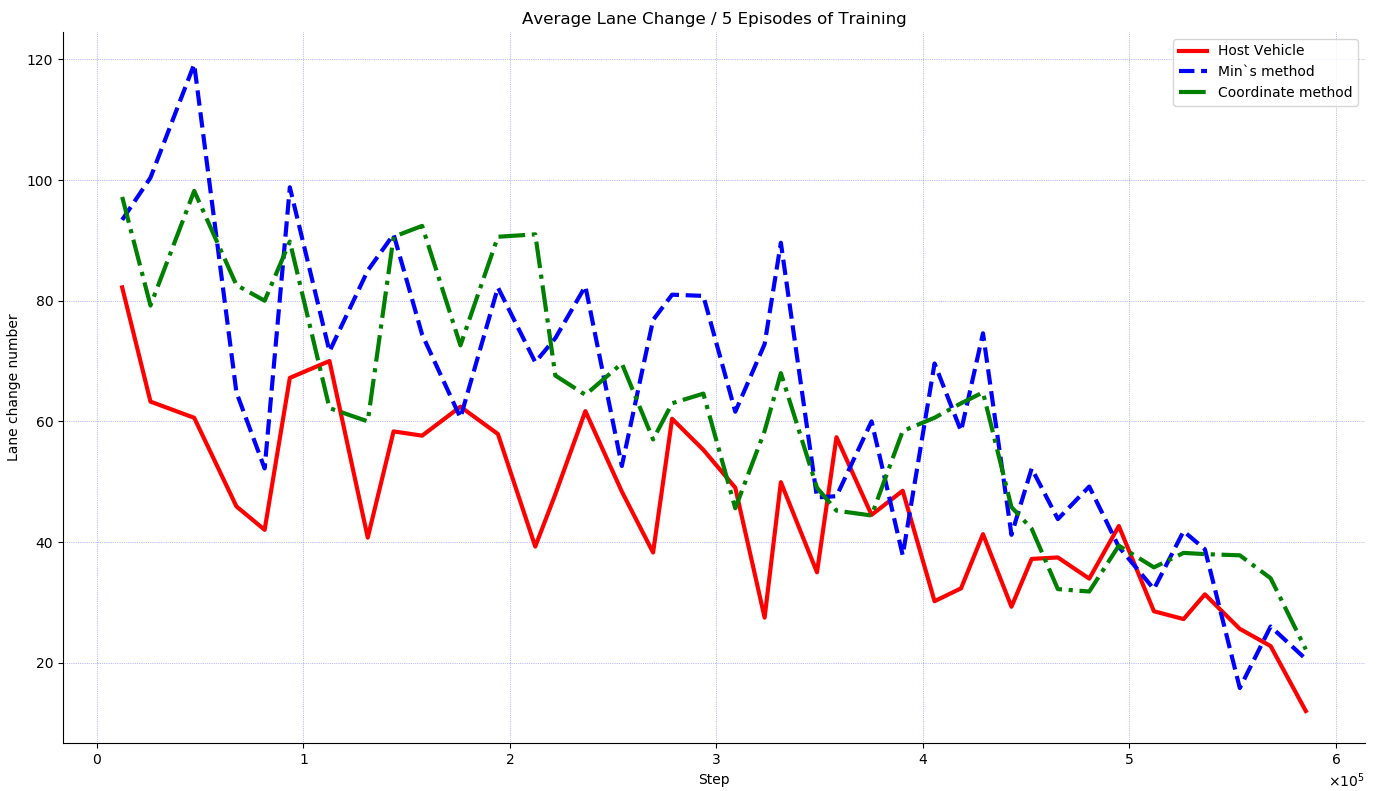}
  \caption{Average lane-change numbers by proposed method, Min's method, Coordinate method.}
  \label{lanechange}
\end{figure}

\begin{figure}[!h]
  \centering
  \includegraphics[width=\hsize]{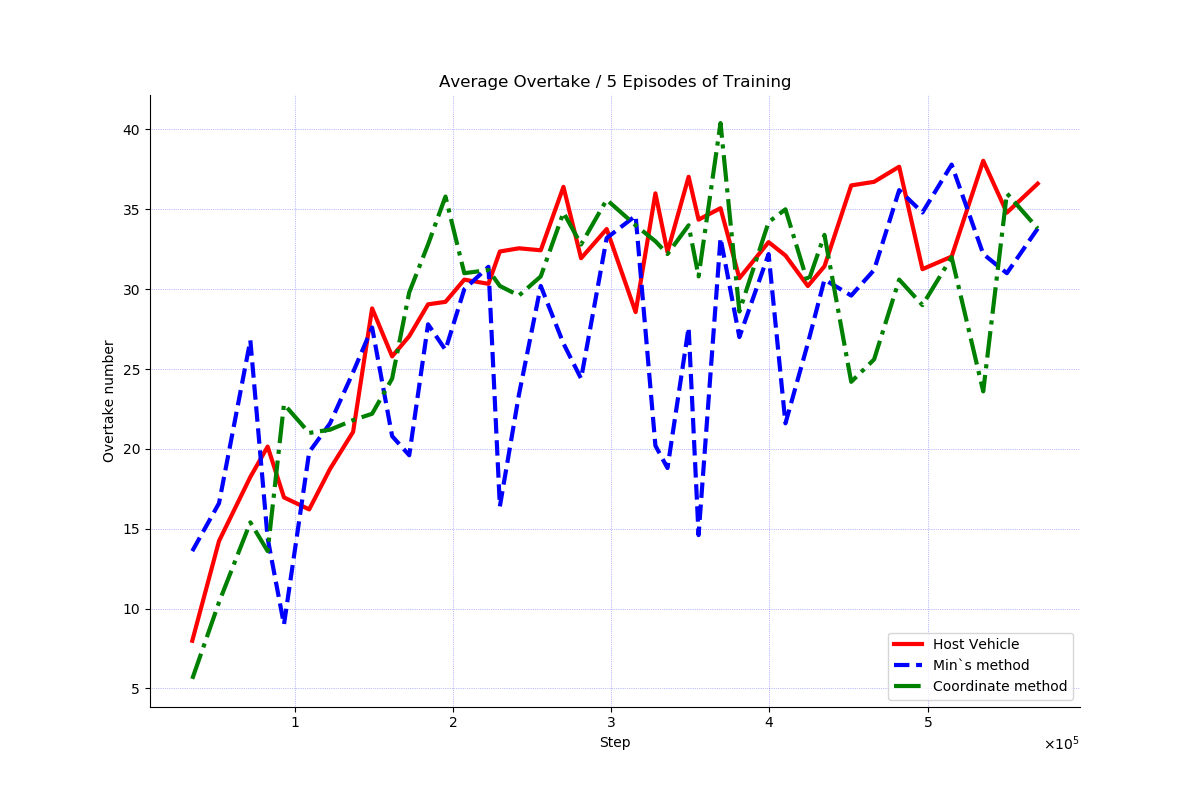}
  \caption{Average overtaking numbers by proposed method, Min's method, Coordinate method.}
  \label{overtake}
\end{figure}

After the training and testing proceeds, all the results are shown as Fig.\ref{speedComp}, Fig.\ref{lanechange}, Fig.\ref{overtake}, and Table~\ref{scenarioresults}.
Firstly, the average speed performance was demonstrated. In Fig.\ref{speedComp}, there are four curves which represent the performance of the proposed model, Min’s model, the Coordinate model, and the non-intelligent model separately. It is obvious that all three learning based methods could make the host vehicle learn the policy to drive faster than the non-intelligent vehicle, but the proposed model has faster training speed and better training results in driving speed.
	
Secondly, Fig.\ref{lanechange} shows the performance of lane change number based on these three learning-based models. In this case, although all the method can reduce the lane-change number during training, the lane-change number of proposed method decrease drastically which means the proposed model has a higher efficiency in learning the lane-change policy. During the testing procedure, the average lane-change number of the proposed model is also less than the others, which is shown in Table.
	
Thirdly, the performance of the overtaking number is shown in Fig.\ref{overtake}. During the training procedure, Min’s method has higher fluctuation in overtaking number while the proposed method and Coordinate method has less fluctuation.

The testing results are shown in Table~\ref{scenarioresults}. There are five testing simulation heterogeneous traffic scenarios which are defined in Table~\ref{scenario}. In considering of the testing results, it is obvious that the proposed method has better generalization capability in coping with heterogeneous traffic flow because the performance is related to the CAV ratio. On the contrary, Min’s method, which based on the raw image as input only, has nearly the same appearances in these cases. For the Coordinate method, the result shows that the performance will decrease without HGM, i.e.,  the capacity of the HGM is verified in this case. All the training and testing above demonstrated that the proposed method can successfully generate the optimal high-level driving behavior for CAV in dense heterogeneous traffic situations to achieve an optimal driving policy and has better generalization ability in coping with different heterogeneous traffic scenarios.
\begin{table}[!htb]
\centering
\caption{Test results of heterogeneous traffic scenarios}
\label{scenarioresults}
\begin{tabular}{l|l|l|l|l|l}
\hhline
			Scenario NO. 	& 1 & 2 & 3 & 4 & 5\\ \hline
			Proposed		&1927&1937&1950&1974&1969 \\
			Min			&1911&1893&1901&1907&1903 \\
			Coordinate		&1868&1893&1867&1875&1907 \\

\hhline
\end{tabular}
\end{table}

\section{Conclusion}

In this paper, we proposed a deep RL based high-level driving behavior decision-making model to learn the optimal driving policy for CV in heterogeneous dense traffic.  Through the HGM mapping process, the raw V2X data is transformed into a unified data format which improves the model performance. Moreover, the proposed deep RL neural network, which combines a two-stream parameter-sharing neural network and Dueling DQN network, successfully learns the optimal high-level driving policy which is driving fast without unnecessary lane changes.  Furthermore, the results of this study show that the proposed model has not only better performance in learning efficiency and robustness but also has significant generalization capability in different heterogeneous traffic scenarios.

\end{document}